\documentclass[10pt,twocolumn,letterpaper]{article}

\usepackage{cvpr}
\usepackage{times}
\usepackage{epsfig}
\usepackage{graphicx}
\usepackage{amsmath}
\usepackage{amssymb}
\usepackage{enumitem}
\usepackage{bm}
\usepackage{booktabs}
\usepackage{subcaption}
\usepackage{multirow}
\DeclareMathOperator*{\argmax}{arg\,max}
\DeclareMathOperator*{\argmin}{arg\,min}

\makeatletter
\newcommand*\bigcdot{\mathpalette\bigcdot@{.7}}
\newcommand*\bigcdot@[2]{\mathbin{\vcenter{\hbox{\scalebox{#2}{$\m@th#1\bullet$}}}}}
\makeatother

\usepackage[breaklinks=true,bookmarks=false]{hyperref}

\cvprfinalcopy 

\def\cvprPaperID{****} 

\begin{document}

\title{Learning to Structure an Image with Few Colors}

\author{Yunzhong Hou~ ~  Liang Zheng~ ~ Stephen Gould\\
	 Australian National University \\
   {\tt\small ~~~\{firstname.lastname\}@anu.edu.au}
}
\maketitle

\begin{abstract}

%
Color and structure are the two pillars that construct an image. Usually, the structure is well expressed through a rich spectrum of colors, allowing objects in an image to be recognized by neural networks. However, under extreme limitations of color space, the structure tends to vanish, and thus a neural network might fail to understand the image. Interested in exploring this interplay between color and structure, we study the scientific problem of identifying and preserving the most informative image structures while constraining the color space to just a few bits, such that the resulting image can be recognized with possibly high accuracy. 
To this end, we propose a color quantization network, ColorCNN, which learns to structure the images from the classification loss in an end-to-end manner. 
Given a color space size, ColorCNN quantizes colors in the original image by generating a color index map and an RGB color palette. Then, this color-quantized image is fed to a pre-trained task network to evaluate its performance. 
In our experiment, with only a 1-bit color space (i.e., two colors), the proposed network achieves 82.1\% top-1 accuracy on the CIFAR10 dataset, outperforming traditional color quantization methods by a large margin. 
For applications, when encoded with PNG, the proposed color quantization shows superiority over other image compression methods in the extremely low bit-rate regime. 
The code is available at \url{https://github.com/hou-yz/color\_distillation}.

\end{abstract}


\section{Introduction}\label{sec:Introduction}
Color and structure are two important aspects of a natural image. The structure is viewed as a combination of shapes, textures, \emph{etc}, and is closely related to colors. Particularly, the structure is well presented only when there exists a sufficient set of colors. 
In this paper, we are interested in how structure can be best presented under color constraints. 

In literature, a closely related line of works is color quantization. Color quantization investigates how to preserve visual similarity in a restricted color space~\cite{heckbert1982color,orchard1991color}. This problem is \textit{human-centered}, as it usually focuses on the visual quality for human viewing.
Particularly, most existing methods are designed under a relatively large color space, \eg, 8-bit, so that quantized images are still visually similar to the original images. In smaller color spaces, \eg, 2-bit or 1-bit, color quantization remains an open question. 




\begin{figure}
    \begin{subfigure}[b]{0.24\linewidth}
    \centering
        \includegraphics[width=\textwidth]{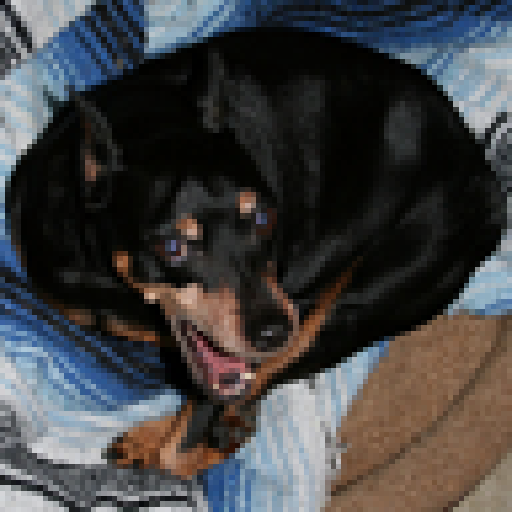}
    \end{subfigure}
    \hfill
    \begin{subfigure}[b]{0.24\linewidth}
    \centering
        \includegraphics[width=\textwidth]{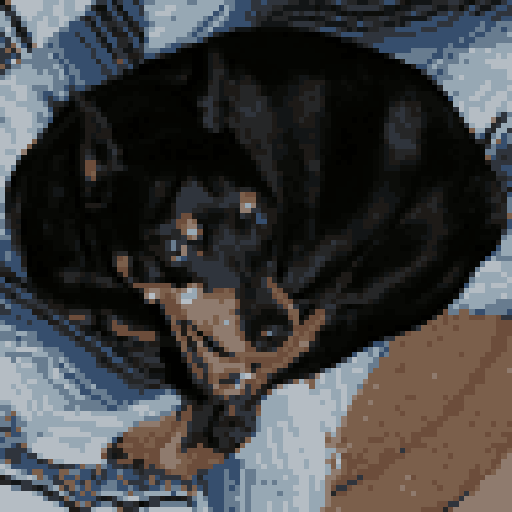}
    \end{subfigure}
    \hfill
    \begin{subfigure}[b]{0.24\linewidth}
    \centering
        \includegraphics[width=\textwidth]{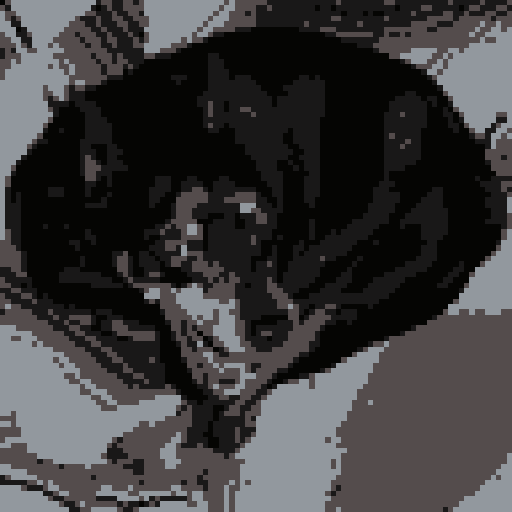}
    \end{subfigure}
    \hfill
    \begin{subfigure}[b]{0.24\linewidth}
    \centering
        \includegraphics[width=\textwidth]{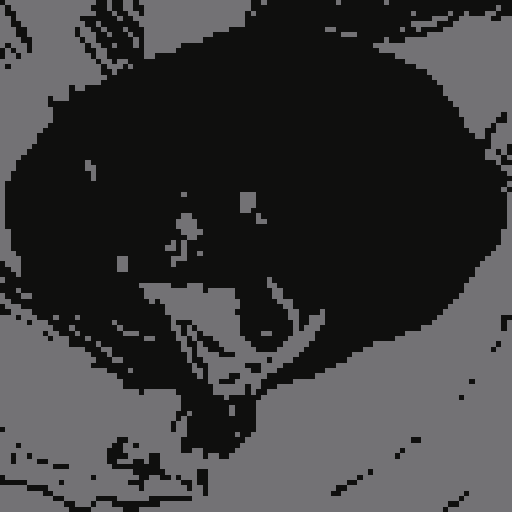}
    \end{subfigure}
    \hfill

    \begin{subfigure}[b]{0.24\linewidth}
    \centering
        \includegraphics[width=\textwidth]{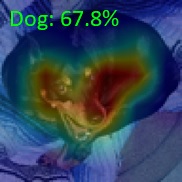}
        \caption{Original}
    \end{subfigure}
    \hfill
    \begin{subfigure}[b]{0.24\linewidth}
    \centering
        \includegraphics[width=\textwidth]{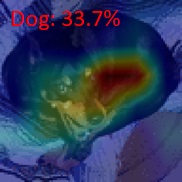}
        \caption{4-bit}
    \end{subfigure}
    \hfill
    \begin{subfigure}[b]{0.24\linewidth}
    \centering
        \includegraphics[width=\textwidth]{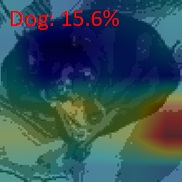}
        \caption{2-bit}
    \end{subfigure}
    \hfill
    \begin{subfigure}[b]{0.24\linewidth}
    \centering
        \includegraphics[width=\textwidth]{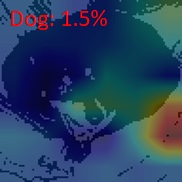}
        \caption{1-bit}
    \end{subfigure}
    \hfill
\vspace{-2mm}
\caption{
Color quantized images (top row) and class activation maps~\cite{zhou2016learning} with softmax probabilities for the ground truth classes (bottom row) in decreasing color space sizes. In ``Original'' (a), colors are described by 24 bits. In (b), (c) and (d), fewer bits are used. We use MedianCut~\cite{heckbert1982color} as the quantization method. When the color space is reduced, the focus of neural network deviates from the informative parts in (a) (correctly recognized (green)), resulting in \textit{recognition failures} (red) in (b), (c), and (d). 
}

\vspace{-4mm}
\label{fig:traditional}
\end{figure}

Natural images usually contain rich colors and structures. When the color space is limited, their connection will compromise the structure. For example, in the first row of Fig.~\ref{fig:traditional}, structures vanish as the color space reduces. 
Moreover, we argue that the neural network trained on original natural images might be ineffective in recognizing quantized images with few colors. In fact, quantized colors and compromised structures drift the attention of the neural network, which might result in recognition failures. For example, in the second row of Fig.~\ref{fig:traditional}, a neural network trained on original images finds the head and body most critical for recognizing the dog. When the color space is reduced gradually,  
the neural network first fails to attend to the head, and then the body, thus leading to recognition failures.

\begin{figure*}
\centering
\includegraphics[width=0.95\linewidth]{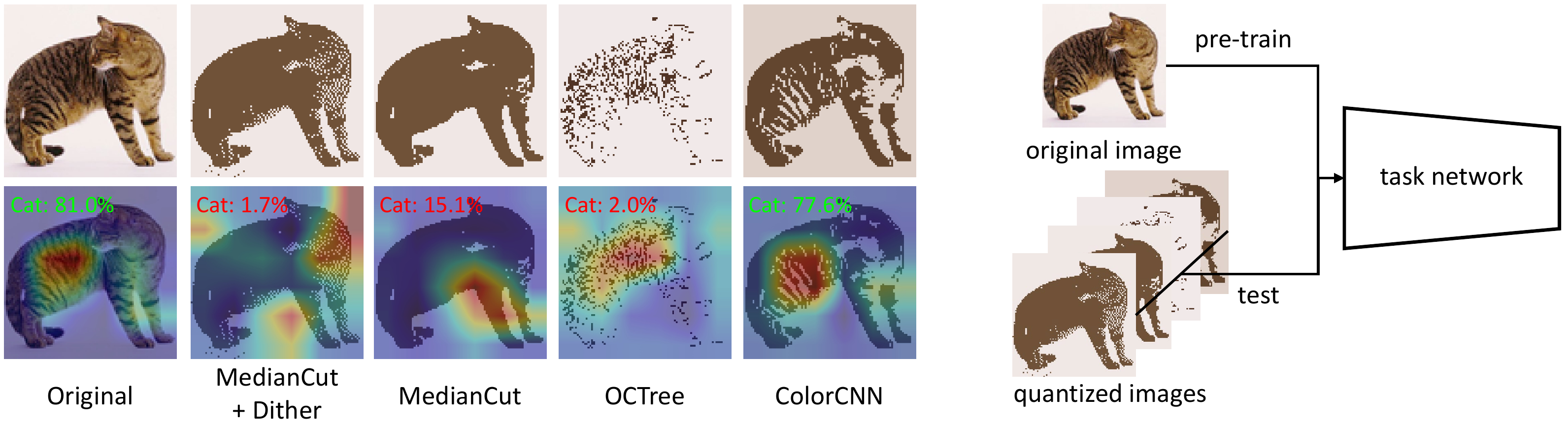}
\vspace{-2mm}
\caption{1-bit color quantization results and evaluation. 
\textbf{Quantization results:} 
``MedianCut + Dither''~\cite{heckbert1982color,floyd1976adaptive}, ``MedianCut''~\cite{heckbert1982color}, and ``OCTree''~\cite{gervautz1988simple} are traditional color quantization methods. ``ColorCNN'' is the proposed method. 
\textbf{Evaluation:} 
We first pre-train a classification network on the original images. Then, we use this network to evaluate the quantized images. 
Traditional methods quantize the original image only based on the color, and hence may lose important shapes and textures. 
When feeding the traditional method results, the pre-trained classifier fails to attend to the informative parts, and make mistakes. 
In comparison, by learning to structure, the image quantized by our method keeps a most similar activation map and thus is successfully recognized by the pre-trained classifier. 
}
\vspace{-4mm}
\label{fig:pre-train_test}
\end{figure*}

In this work, we study a scientific problem: how to preserve the critical structures under an extremely small color space? 
This problem is orthogonal to traditional color quantization problems, as it is \textit{task-centered}: neural network recognition accuracy is its major focus, instead of human viewing. 
As shown in Fig.~\ref{fig:pre-train_test}, we evaluate the (quantized) images with a classifier pre-trained on original images. 
To optimize the recognition accuracy, we design a color quantization method, ColorCNN, which learns to structure the image in an end-to-end manner. 
Unlike traditional color quantization methods that solely rely on color values for the decision, ColorCNN exploits colors, structures, and semantics to spot and preserve the critical structures. It quantizes the image by computing a color index map and assigning color palette values. 
In Fig.~\ref{fig:pre-train_test}, images quantized by ColorCNN enables the classifier to successfully focus on the cat tabby and forelimb. In this example, attending to the informative regions leads to correct recognition. 

We demonstrate the effectiveness of ColorCNN quantization on classification tasks. With a few colors, we show that the proposed methods outperforms traditional ones by a large margin. Four datasets, including 
CIFAR10~\cite{krizhevsky2009learning}, CIFAR100~\cite{krizhevsky2009learning}, STL10~\cite{coates2011analysis}, and tiny-imagenet-200~\cite{le2015tiny}, and three classification networks, including AlexNet~\cite{krizhevsky2012imagenet}, VGG~\cite{simonyan2014very}, and ResNet~\cite{he2016deep}, are used for testing.
For applications, the ColorCNN image quantization can be used in extremely low bit-rate image compression. 



\section{Related Work}\label{sec:Related_Work}
\textbf{Color quantization.} 
Color quantization~\cite{orchard1991color,deng1999peer,achanta2012slic,deng2001unsupervised,wu1992color} clusters the colors to reduce the color space while keeping visual similarity. Heckbert \etal~\cite{heckbert1982color} propose the popular MedianCut method. Later, Gervautz \etal~\cite{gervautz1988simple} design another commonly-used color quantization method, OCTree.
Dithering~\cite{floyd1976adaptive}, which removes visual artifacts by adding a noise pattern, 
is also studied as an optional step. 
The color quantized images can be represented as \textit{indexed color}~\cite{poynton2012digital}, and encoded with PNG~\cite{boutell1997png}.

\textbf{Human-centered image compression.} Based on heuristics, many image compression methods are designed for human viewers. 
These methods fall into two categories, lossless compression, \eg, PNG~\cite{boutell1997png}, and lossy compression, \eg, JPEG~\cite{wallace1992jpeg,skodras2001jpeg} and \textit{color quantization}. 

Recently, deep learning methods are introduced to image compression problems. Both recurrent methods~\cite{oord2016pixel,johnston2018improved,toderici2017full,toderici2015variable} and convolutional methods~\cite{mentzer2019practical,li2018learning,van2016conditional,agustsson2018generative,agustsson2017soft,theis2017lossy} are investigated.
Ballé \etal~\cite{balle2016end} propose generalized divisive normalization (GDN) for image compression. Agustsson \etal~\cite{agustsson2018generative} propose a multi-scale discriminator in a generative adversarial network (GAN) for low-bitrate compression. 
In ~\cite{liu2018feature,jia2019comdefend}, researchers apply image compression methods to defend against adversarial attacks. 

\textbf{Task-centered image compression.} 
Traditional or deep-learning based, the fore-mentioned image compression methods are human-centered. 
Liu \etal~\cite{liu2019machine} points out that for segmentation, human-centered compression is not the best choice for 3D medical images. 
For 2D map data and 3D scene models, task-centered compression methods are designed for localization~\cite{wei2019learning,camposeco2019hybrid}.
%

\textbf{Neural network recognition with compressed data.} Some researchers work on certain tasks with compressed data. For example, solving action recognition problem with compressed video~\cite{wu2018compressed,zhang2016real,zhang2018real,shou2019dmc}. Wang \etal~\cite{wang2019fast} accelerate video object recognition by exploiting compressed data.

\begin{figure*}[t]
\begin{center}
\centering
\includegraphics[width=0.95\linewidth]{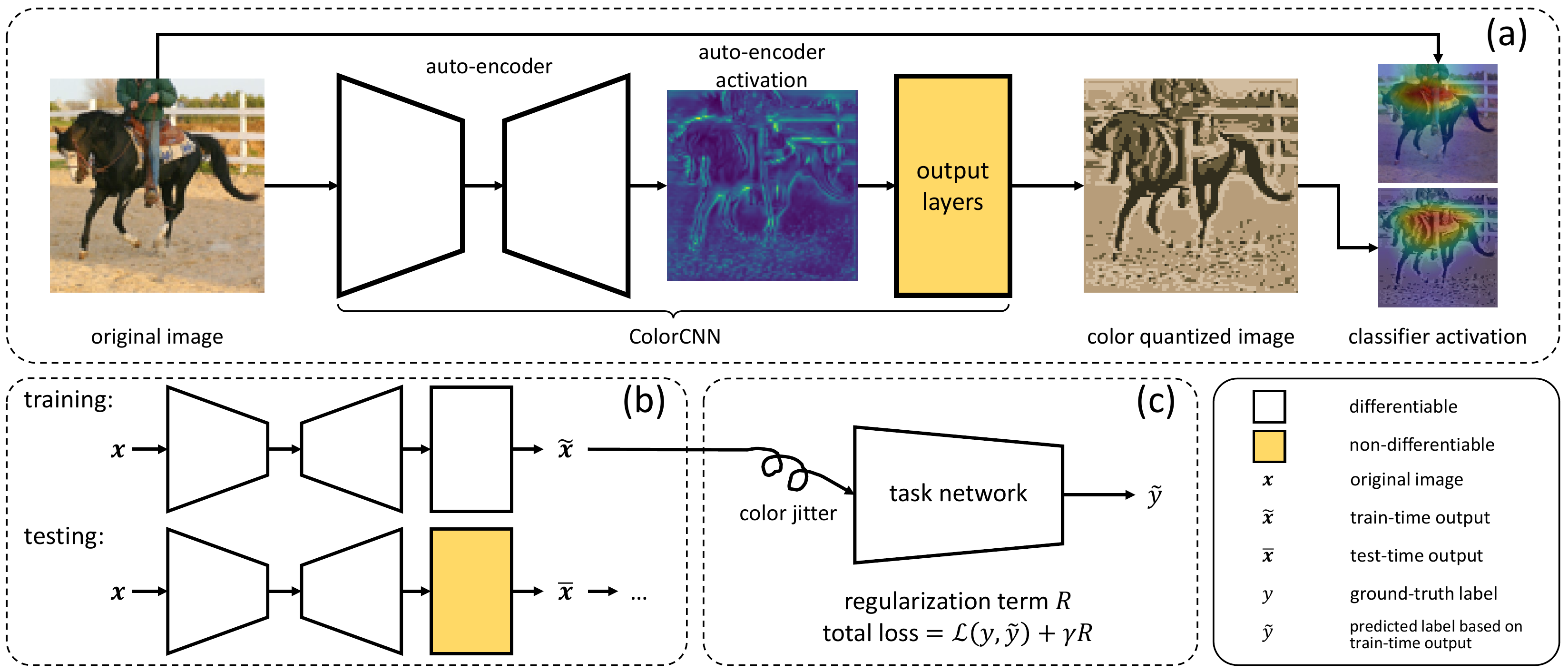}
\end{center}
\vspace{-4mm}
\caption{Overview of our color quantization method. (\textbf{a}): ColorCNN can identify (auto-encoder activation) and preserve (quantized image) the critical structures in the original image. Its output has a similar class activation map to the original image. 
(\textbf{b}): we replace the non-differentiable parts with approximations during training. (\textbf{c}): one regularization term is introduced to keep the approximation similar to the original network. Also, we add a color jitter to the quantized image to prevent premature convergence. The ColorCNN network is trained with classification loss in an end-to-end manner. 
}
\vspace{-3mm}
\label{fig:system}
\end{figure*}

\section{Motivation}\label{sec:recognition_failure}
In an extremely small color space, we find the traditional color quantization methods fail to preserve the critical structures. This is because these methods usually take the color-value-only approach to cluster the colors, completely ignoring structures. 
However, as some~\cite{geirhos2018imagenet,Goodfellow-et-al-2016} suggest, critical shapes and textures with semantic meaning play an important role in neural network recognition. In fact, we witness attention drifts when the structure is not well preserved in Fig.~\ref{fig:traditional} and Fig.~\ref{fig:pre-train_test}. This further leads to recognition failure. 
Motivated by this failure, in the following sections, we further investigate how to effectively preserve the structures in an extremely small color space.


\section{Proposed Approach}\label{sec:Method}

To identify and preserve the critical structures in the original image, we design ColorCNN (see Fig.~\ref{fig:system}). In this section, we first formulate the learning-to-structure problem mathematically. Next, we present ColorCNN architecture. At last, we provide an end-to-end training method.

\subsection{Problem Formulation}\label{sec:Problem_formulation}

Without loss of generality, we can define a classification network with parameter $\theta$ as $f_\theta\!\left(\cdot\right)$. For an image label pair $\left(\bm{x},y\right)$, this network estimate its label $\hat{y}=f_\theta\!\left(\bm{x}\right)$. To train the network, we minimize its loss $\mathcal{L}\left(y, \hat{y} \right)$ on dataset $D$, 
\begin{equation}\label{eq:classifier_general}
    \theta^\star = \argmin_{\theta}{\sum_{\left(\bm{x},y\right)\in D}{\mathcal{L}\left(y, f_\theta\!\left(\bm{x}\right)\right)}},
\end{equation}
where $\theta^\star$ denotes the optimal parameter.

For color quantization, we design ColorCNN architecture. Given input image $\bm{x}$, it can output the color quantized image $\bm{\overline{x}}$, by computing the color index map $M\left(\bm{x}\right)$ and the color palette $T\left(\bm{x}\right)$. 
We represent the forward pass for ColorCNN as one function $\bm{\overline{x}}=g_\psi\!\left(\bm{x}\right)$ with parameter $\psi$.


Our objective is to structure an image with few colors such that a pre-trained classifier has a possibly high accuracy on the color quantized images. It is written as 
\begin{equation}\label{eq:optim_quantizer}
\psi^\star = \argmin_{\psi}{\sum_{\left(\bm{x},y\right)\in D}{\mathcal{L}\left(y, f_{\theta^\star}\!\left(g_\psi\!\left(\bm{x}\right)\right)\right)+\gamma R}},
\end{equation}
where $\psi^\star$ denotes the optimized parameter for ColorCNN, given the pre-trained classifier parameterized by $\theta^\star$. $R$ is a regularization term and $\gamma$ denotes its weight.

\begin{figure}[t]
\begin{center}
\centering
\includegraphics[width=\linewidth]{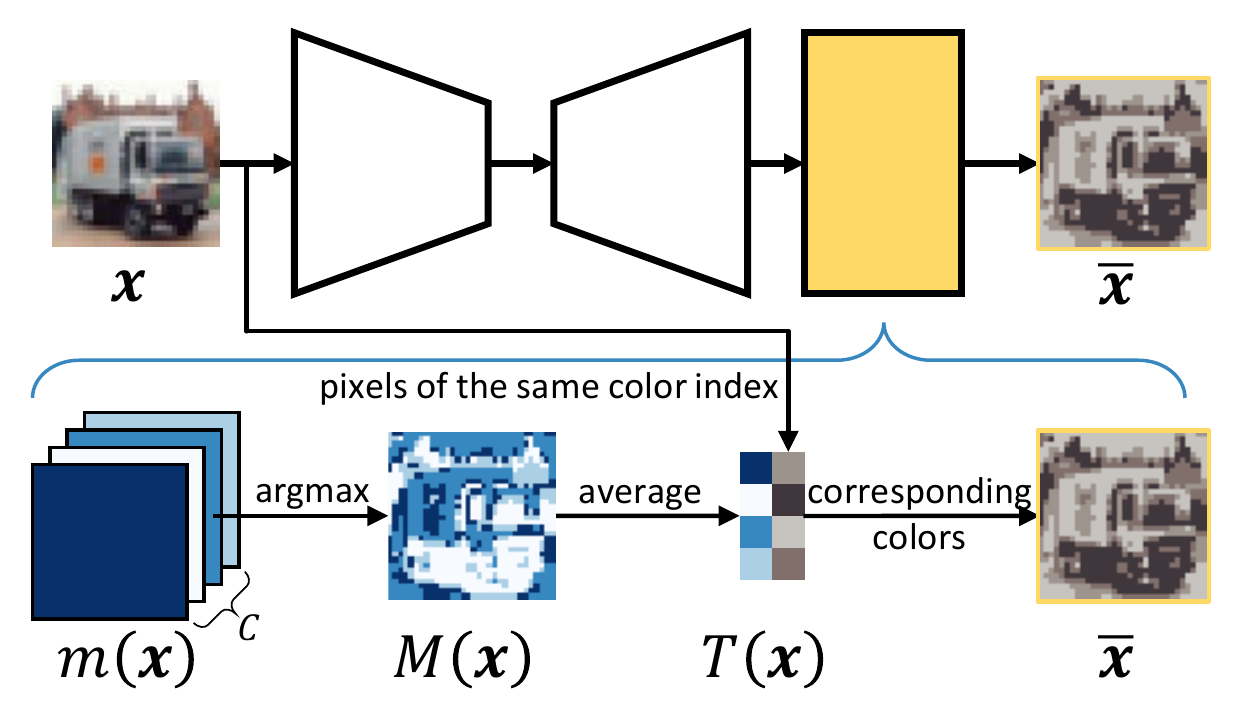}
\vspace{-3mm}
\caption{ColorCNN architecture (\textbf{test-time}). First, the convolutional layers output a $C$-channel probability map $m\left(\bm{x}\right)$ for $C$ colors. Next, a $1$-channel color index map $M\left(\bm{x}\right)$ is created via the $\argmax$ function. Then, the color palette $T\left(\bm{x}\right)$ is computed as average of all pixels that are of the \textit{same color index}. At last, the color quantized image $\bm{\overline{x}}$ is created via a \textit{table look-up} session. 
}
\vspace{-3mm}
\label{fig:forward_test}
\end{center}
\end{figure}

\subsection{ColorCNN Architecture}\label{sec:ColorCNN_structure}
We show the ColorCNN architecture in Fig.~\ref{fig:forward_test}. 
Its first component is an U-net~\cite{ronneberger2015u} auto-encoder that identifies the critical and semantic-rich structures (auto-encoder activation map in Fig.~\ref{fig:system}).


For the second component, two depth-wise ($1\times1$ kernel size) convolution layers create a softmax probability map of each pixel taking one specific color. 
This results in a $C$-channel probability map $m\left(\bm{x}\right)$ (softmax over $C$-channel). 

Then, for each input image $\bm{x}$, the $1$-channel color index map $M\left(\bm{x}\right)$ is computed as the $\argmax$ over the $C$-channel probability map $m\left(\bm{x}\right)$,
\begin{equation}\label{eq:color_index_test}
    M\left(\bm{x}\right) = \argmax_{c}{m\left(\bm{x}\right)}.
\end{equation}

The RGB color palette, $T\left(\bm{x}\right)$, which is of shape $C\times 3$, is computed as average of all pixels that falls into certain quantized color index,
\begin{equation}\label{eq:color_palette_test}
    \left[T\left(\bm{x}\right)\right]_c = \frac{\sum_{\left(u,v\right)}{\left[\bm{x}\right]_{u,v} \bigcdot \mathbb{I}\left(\left[M\left(\bm{x}\right)\right]_{u,v}=c\right)}}{\sum_{\left(u,v\right)}{\mathbb{I}\left(\left[M\left(\bm{x}\right)\right]_{u,v}=c\right)}},
\end{equation}
where $\left[\cdot\right]_i$ denotes the element or tensor with index $i$. $\mathbb{I}\left(\cdot\right)$ is an indicator function. 
$\bigcdot$ denotes the pointwise multiplication. 
For pixel $\left(u,v\right)$ in a $W\times H$ image, $\left[x\right]_{u,v}$ denotes the pixel and its RGB value in the input image, and $\left[M\left(\bm{x}\right)\right]_{u,v}$ represents its computed color index. $\left[T\left(\bm{x}\right)\right]_c$ denotes the RGB value for the quantized color $c$. 

At last, the quantized image $\bm{\overline{x}}$ is created as
\begin{equation}\label{eq:reconsturect_test}
    \bm{\overline{x}} = \sum_{c}{\left[T\left(\bm{x}\right)\right]_c\bigcdot \mathbb{I}\left(M\left(\bm{x}\right)=c\right)}.
\end{equation}

By combining Eq.~\ref{eq:color_index_test},~\ref{eq:color_palette_test},~\ref{eq:reconsturect_test}, we finish the ColorCNN forward pass $\bm{\overline{x}} = g_\psi\!\left(\bm{x}\right)$. 


\begin{figure}[t]
\begin{center}
\centering
\includegraphics[width=\linewidth]{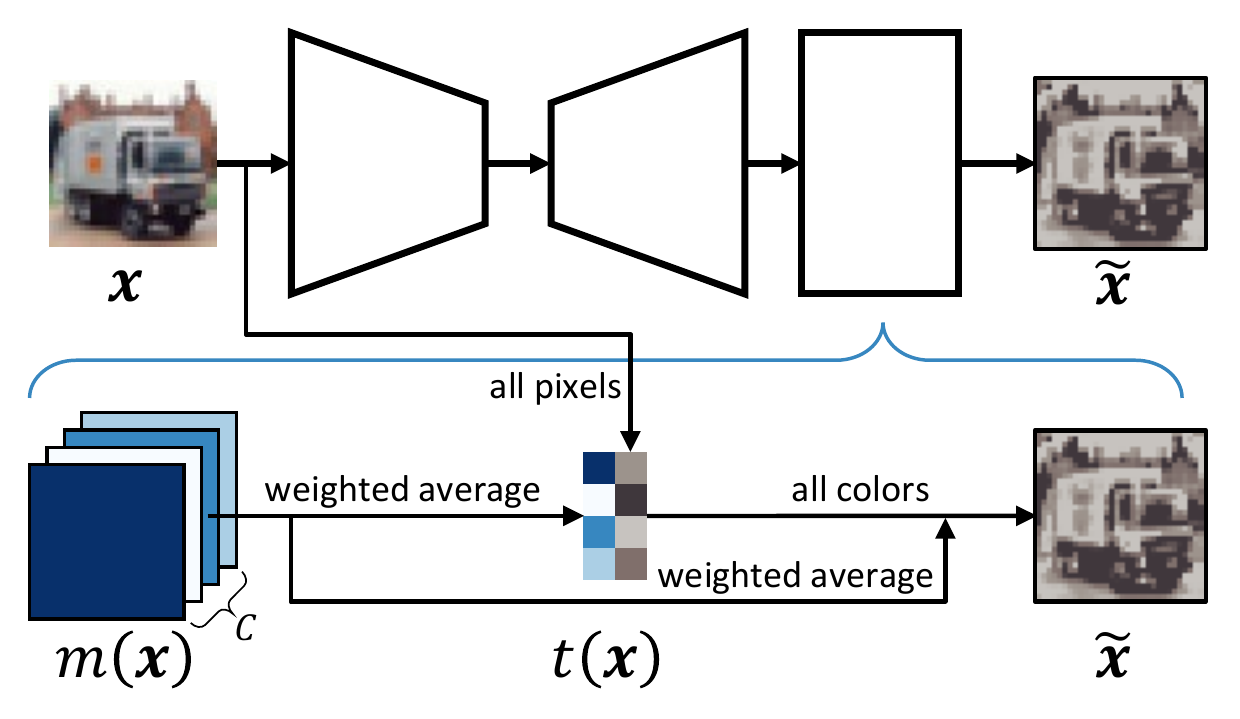}
\end{center}
\vspace{-3mm}
\caption{The differentiable approximation (\textbf{train-time}). The $C$-channel probability map $m\left(\bm{x}\right)$ is used instead of  the $\argmax$ color index map $M\left(\bm{x}\right)$. Next, the color palette $t\left(\bm{x}\right)$ is adjusted as weighted average over \textit{all} pixels. At last, instead of table look-up, the quantized image $\bm{\widetilde{x}}$ is computed as the weighted average of \textit{all} colors in the color palette. 
}
\vspace{-3mm}
\label{fig:forward_train}
\end{figure}

\subsection{End-to-End Learning}

\subsubsection{Differentiable Approximation}
Fig.~\ref{fig:forward_train} shows the differentiable approximation in training. 
To start with, we remove the $\argmax$ $1$-channel color index map $M\left(\bm{x}\right)$ in Eq.~\ref{eq:color_index_test}. Instead, we use the $C$-channel softmax probability map $m\left(\bm{x}\right)$.


Next, we change the color palette design that follows. 
For each quantized color, instead of averaging the pixels of the \textit{same color index}, we set its RGB color value $\left[t\left(\bm{x}\right)\right]_c$ as the weighted average over \textit{all} pixels, 
\begin{equation}\label{eq:color_palette_train}
    \left[t\left(\bm{x}\right)\right]_c = \frac{\sum_{\left(u,v\right)}{\left[\bm{x}\right]_{u,v} \bigcdot \left[m\left(\bm{x}\right)\right]_{u,v,c}}}{\sum_{\left(u,v\right)}{\left[m\left(\bm{x}\right)\right]_{u,v,c}}},
\end{equation}
Here, the $C$-channel probability distribution $\left[m\left(\bm{x}\right)\right]_{u,v}$ is used as the contribution ratio of pixel $\left(u,v\right)$ to $C$ colors. This will result in a slightly different color palette $t\left(\bm{x}\right)$.

In the end, we change the table look-up process from the original forward pass into a weighted sum. For quantized color with index $c$, we use $\left[m\left(\bm{x}\right)\right]_{c}$ as the intensity of expression over entire image. Mathematically, the train-time quantized image $\bm{\widetilde{x}}$ is computed as
\begin{equation}\label{eq:reconsturect_train}
    \bm{\widetilde{x}} = \sum_{c}{\left[t\left(\bm{x}\right)\right]_c\bigcdot \left[m\left(\bm{x}\right)\right]_{c}}.
\end{equation}

By combining Eq.~\ref{eq:color_palette_train},~\ref{eq:reconsturect_train}, the forward pass for ColorCNN during training can be formulated as $\bm{\widetilde{x}} = \widetilde{g_\psi}\!\left(\bm{x}\right)$. At last, we substitute ${g_\psi}\left(\cdot\right)$ with $\widetilde{g_\psi}\left(\cdot\right)$ in Eq.~\ref{eq:optim_quantizer} for end-to-end training.

\begin{figure}[t!]
    \centering
    \begin{subfigure}[b]{0.45\linewidth}
    \centering
        \includegraphics[width=\textwidth]{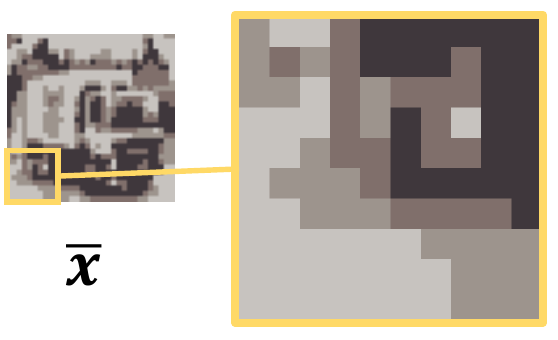}
        \caption{Test-time result}
    \end{subfigure}
    \hfill 
    \begin{subfigure}[b]{0.45\linewidth}
    \centering
        \includegraphics[width=\textwidth]{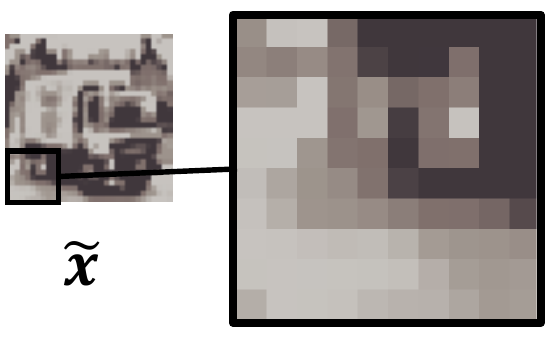}
        \caption{Train-time result}
    \end{subfigure}
\vspace{-2mm}
\caption{Comparison between test-time result $\bm{\overline{x}}$ and train-time result $\bm{\widetilde{x}}$. Each pixel in $\bm{\widetilde{x}}$ is a weighted average of all colors in its palette. Thus, more colors are introduced. 
}
\vspace{-3mm}
\label{fig:comparison}
\end{figure}

Even though the two forward pass ${g_\psi}\left(\cdot\right)$ and $\widetilde{g_\psi}\left(\cdot\right)$ use the same parameter $\psi$, they behave very differently. See Fig.~\ref{fig:comparison} for a side-by-side comparison between the outputs.  
The test-time output $\bm{\overline{x}}$ only has $C$ colors, whereas the train-time output $\bm{\widetilde{x}}$ has more than $C$ colors. 
The main reason for this mismatch boils down to the difference between one-hot and softmax vectors. As shown in Fig.~\ref{fig:forward_test}, the one-hot approach allows influence only from \textit{some} pixels to one quantized color, and from \textit{one} color to any quantized pixel. On the other hand, in Fig.~\ref{fig:forward_train}, with softmax function, \textit{all} pixels influence all colors in the palette, and \textit{all} colors in the palette contribute to each pixel in the output image.

\subsubsection{Regularization}\label{sec:Regularization}
During training, the softmax probability distribution in each pixel can deviate far from a one-hot vector, which easily creates more than $C$ colors. On the other hand, during test-time, it cannot be guaranteed that the quantized image will contain exactly $C$ colors: it is possible for the test-time quantization results to use fewer than $C$ colors. 

In order to make sure all $C$ colors are used for the quantization, we propose a regularization term. It encourages each of the $C$ colors to be selected with very high possibilities by at least one pixel in the image. 
For pixel $\left(u,v\right)$, we maximize the largest value of the probability distribution $\left[m\left(\bm{x}\right)\right]_{u,v}$. The regularization term is designed as
\begin{equation}\label{eq:regularization1}
    R = \log_{2}{C}\times \left(1-\frac{1}{C}\times \sum_{c}{\max_{\left(u,v\right)}{\left[m\left(\bm{x}\right)\right]_{u,v}}}\right).
\end{equation}
We take the minus of the summation, since it is desired to minimize this regularization term $R$. We also offset the regularization term by $1$ in order to make it positive. 



\subsubsection{Color Jitter}

We train the proposed ColorCNN with a pre-trained classifier $f_{\theta^\star}\!\left(\cdot\right)$ (see Eq.~\ref{eq:optim_quantizer}). 
During training, as long as ColorCNN can provide barely satisfactory results, the pre-trained classifier will have a good chance in making the correct decision. 
However, given more freedom in the train-time results, the test-time results might still be struggling when the network converges. 
We refer to this phenomenon as premature convergence. 

In order to prevent this premature convergence, during training, we add a jitter $\xi\times n$ to the color quantized image $\bm{\widetilde{x}}$ after normalization. The noise $n$ is sampled from a Gaussian distribution $\mathcal{N}\left(0,1\right)$. $\xi$ denotes its weight. 
\textbf{First}, higher variance in training output delays convergence (more difficult to fit data with higher variance), allowing a better-trained network.
\textbf{Second}, feature distribution of training output with color jitter (higher variance) may have more overlap with that of the testing output. 
Since the classifier can recognize the color jittered training output, it may perform better during testing. 


\begin{figure}[t]
    \centering
    \begin{subfigure}[b]{0.24\linewidth}
    \centering
        \includegraphics[width=\textwidth]{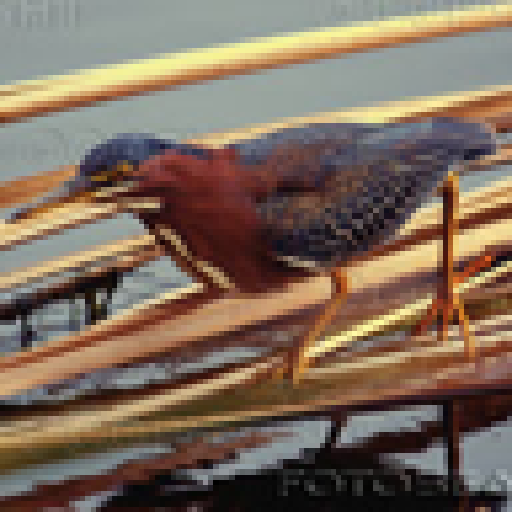}
        \caption{Original}
    \end{subfigure}
    \hfill 
    \begin{subfigure}[b]{0.24\linewidth}
    \centering
        \includegraphics[width=\textwidth]{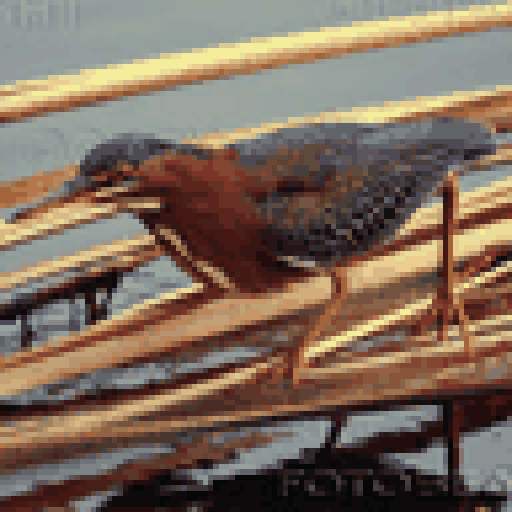}
        \caption{MedianCut}
    \end{subfigure}
    \hfill
    \begin{subfigure}[b]{0.24\linewidth}
    \centering
        \includegraphics[width=\textwidth]{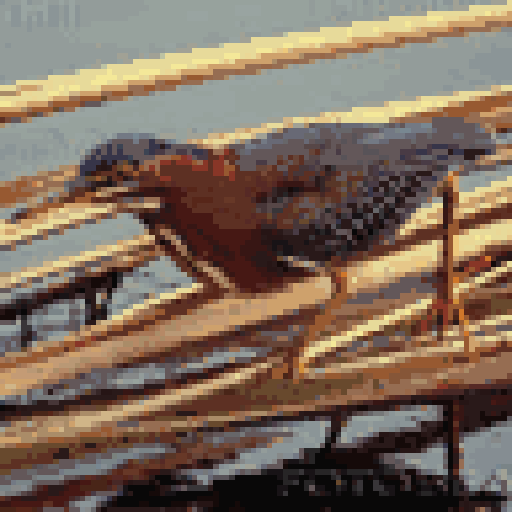}
        \caption{OCTree}
    \end{subfigure}
    \hfill 
    \begin{subfigure}[b]{0.24\linewidth}
    \centering
        \includegraphics[width=\textwidth]{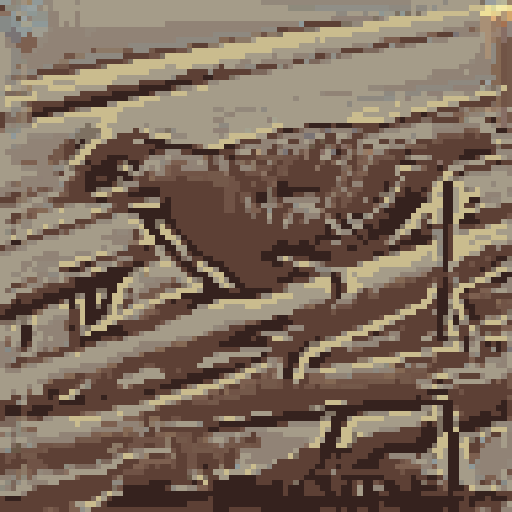}
        \caption{ColorCNN}
    \end{subfigure}
\vspace{-2mm}
\caption{6-bit color quantization results. 
Traditional methods formulate color quantization as a clustering problem. On the contrary, we formulate this problem as per-pixel classification. When a bigger number-of-cluster (color space size $C$) is assumed, classification-based method (ColorCNN) cannot compete with clustering-based methods. For more details, see Section~\ref{sec:Discussion}.
}
\label{fig:6-bit}
\vspace{-3mm}
\end{figure}

\section{Discussion}\label{sec:Discussion}

\textbf{Larger color space is not the focus of this work.} When the color space is larger, the richer colors naturally support more structures, making structure preserving less of a scientific problem. 
Moreover, many traditional methods study this problem, and their quantization results achieve very good accuracy on pre-trained classifiers. In fact, in a 6-bit color space, accuracy of quantized images only fall behind original images marginally (see Section~\ref{sec:Evaluation_of_ColorCNN}). 

\textbf{ColorCNN versus clustering-based color quantization.} 
Using a per-pixel classification formulation, ColorCNN cannot provide a competitive result to clustering-based methods, when the color space is large (Fig.~\ref{fig:6-bit}). 
This is very normal, given that per-pixel classification treats each pixel individually, whereas clustering consider all pixels to enforce intra-cluster similarity to make a collective decision. The feature of each pixel has global info, but this is insufficient: making decisions globally and collectively is needed. 
With a competent end-to-end neural network clustering method, ColorCNN can possibly out-perform traditional methods, even in a large color space. 
Why not incorporating clustering in ColorCNN, one may ask. 
In fact, using neural networks to solve the clustering problem is non-trivial, and stands as a challenging problem itself. Some pioneer works in neural network clustering investigate end-to-end feature update, or feature dimension reduction~\cite{xie2016unsupervised,aljalbout2018clustering}. Still, they must rely on k-means or other clustering methods during testing. 
In fact, neural network clustering itself is a different line of work, and is beyond the scope of this paper.

\section{Experiment}
\subsection{Experiment Setup}\label{sec:Experiment_Setup}

\textbf{Datasets.} We test  ColorCNN performance on multiple datasets. CIFAR10 and CIFAR100 datasets~\cite{krizhevsky2009learning} include 10 and 100 classes of general objects, respectively. 
Similar to CIFAR10, STL10 dataset~\cite{coates2011analysis} also contains 10 image classes. However, the images in STL10 have a higher $96\times 96$ resolution. 
We also evaluate on the tiny-imagenet-200 dataset~\cite{le2015tiny}, which is a subset of ImageNet dataset~\cite{deng2009imagenet}. It has 200 classes of $64\times 64$ general object images. We compare the four datasets in Table~\ref{tab:datasets}.

\textbf{Evaluation.} For evaluation, we report top-1 classification accuracy on the mentioned datasets. 


\textbf{Classification networks.} We choose AlexNet~\cite{krizhevsky2012imagenet}, VGG16~\cite{simonyan2014very} and ResNet18~\cite{he2016deep} for classification network. 
All classifier networks are trained for $60$ epochs with a batch size of $128$, except for STL10, where we set the batch size to $32$. We use an SGD optimizer with a momentum of $0.5$, L2-normalization of $5\times 10^{-4}$. We choose the 1cycle learning rate scheduler~\cite{smith2019super} with a peak learning rate at $0.1$.

\begin{table}[t!]
\centering
\resizebox{\linewidth}{!}{
\begin{tabular}{l|c|c|c|c}
\toprule
             & \multicolumn{1}{c|}{CIFAR10} & \multicolumn{1}{c|}{CIFAR100} & \multicolumn{1}{c|}{STL10} & \multicolumn{1}{c}{Tiny200} \\ \hline
\#class      & 10                           & 100                           & 10                         & 200                                    \\ \hline
Train images & 50,000                       & 50,000                        & 5,000                      & 100,000                                \\ \hline
Test images  & 10,000                       & 10,000                        & 8,000                      & 10,000 $^\dag$                         \\ \hline
Resolution   & $32\times32$                 & $32\times32$                  & $96\times96$               & $64\times64$                           \\ \bottomrule
\end{tabular}
}
\vspace{-1mm}
\caption{Datasets comparison. ``Tiny200'' denotes tiny-imagenet-200 dataset. $\dag$ indicates we use validation set for testing, given that the online test set is not available. }
\vspace{-3mm}
\label{tab:datasets}
\end{table}

\textbf{ColorCNN.} We train ColorCNN on top of the original image pre-trained classifier. We set the hyper-parameters as follows. For the regularization and color jitter weight, we set $\gamma=1$ and $\xi=1$. We also normalize the quantized image by $4\times$ the default variance of the original images, so as to prevent premature convergence. 
For training, we run the gradient descent for 60 epochs with a batch size of $128$. Similar to classification networks, we also reduce the batch size to $32$ on the STL10 dataset. The SGD optimizer for ColorCNN training is the same as for classifier networks. For the learning rate scheduler, we choose Cosine-Warm-Restart~\cite{loshchilov2016sgdr} with a peak learning rate at $0.01$, minimal learning rate at $0$, and uniform restart period of $20$.

We finish all the experiment on one RTX-2080TI GPU.

\subsection{Evaluation of ColorCNN}\label{sec:Evaluation_of_ColorCNN}

\textbf{Classification network performance.} We report the top-1 test accuracy of the classification networks in Table~\ref{tab:classifier}. There is a consistent accuracy increase going from AlexNet to VGG16 to ResNet18 on all four datasets.

\begin{figure}
\centering
    \begin{subfigure}[b]{0.24\linewidth}
    \centering
        \includegraphics[width=\textwidth]{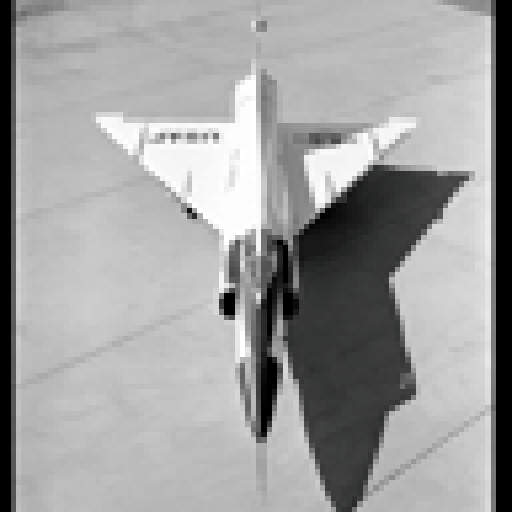}
    \end{subfigure}
    \begin{subfigure}[b]{0.24\linewidth}
    \centering
        \includegraphics[width=\textwidth]{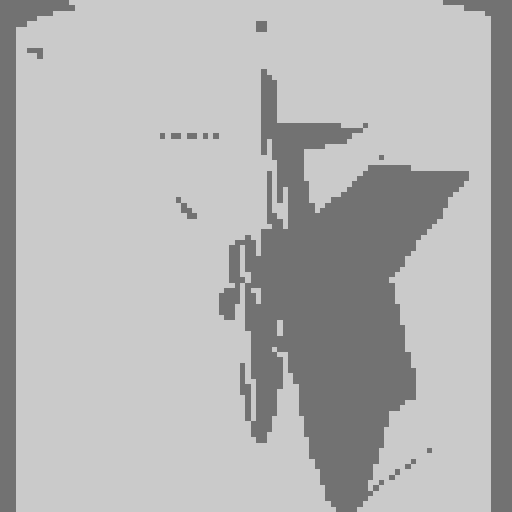}
    \end{subfigure}
    \begin{subfigure}[b]{0.24\linewidth}
    \centering
        \includegraphics[width=\textwidth]{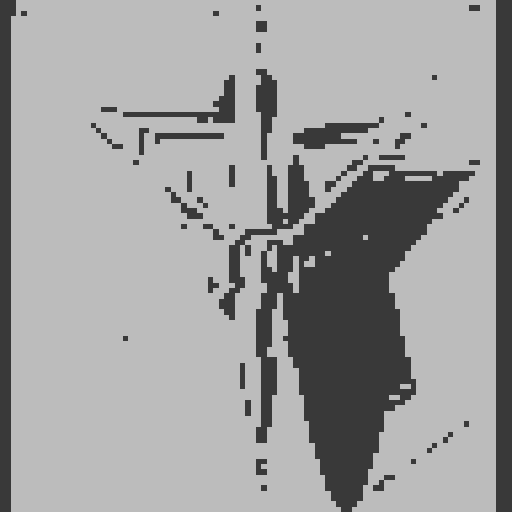}
    \end{subfigure}
    \begin{subfigure}[b]{0.24\linewidth}
    \centering
        \includegraphics[width=\textwidth]{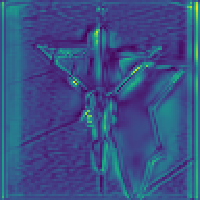}
    \end{subfigure}
    \hfill
    
    \begin{subfigure}[b]{0.24\linewidth}
    \centering
        \includegraphics[width=\textwidth]{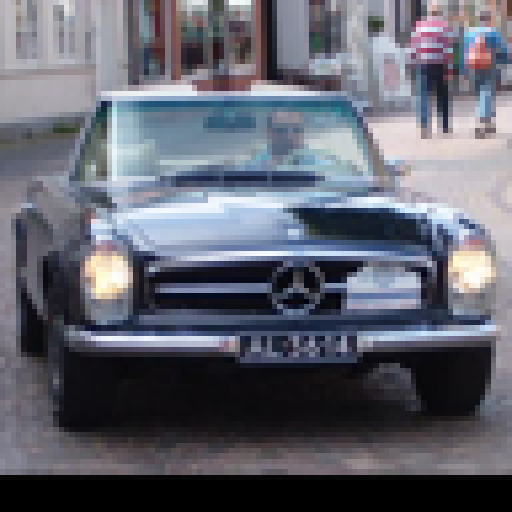}
    \end{subfigure}
    \begin{subfigure}[b]{0.24\linewidth}
    \centering
        \includegraphics[width=\textwidth]{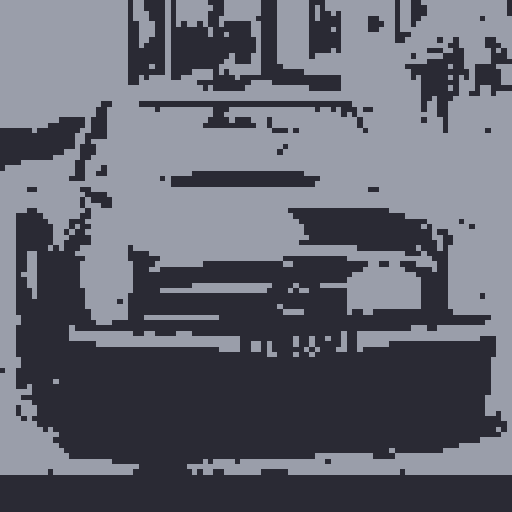}
    \end{subfigure}
    \begin{subfigure}[b]{0.24\linewidth}
    \centering
        \includegraphics[width=\textwidth]{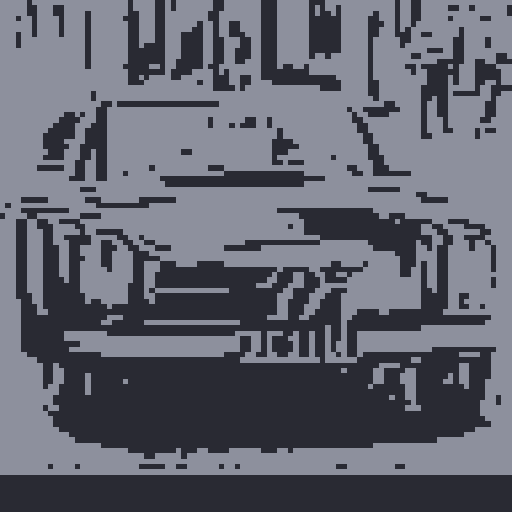}
    \end{subfigure}
    \begin{subfigure}[b]{0.24\linewidth}
    \centering
        \includegraphics[width=\textwidth]{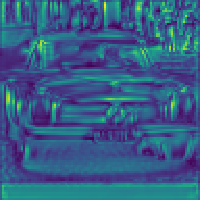}
    \end{subfigure}
    \hfill
    
    \begin{subfigure}[b]{0.24\linewidth}
    \centering
        \includegraphics[width=\textwidth]{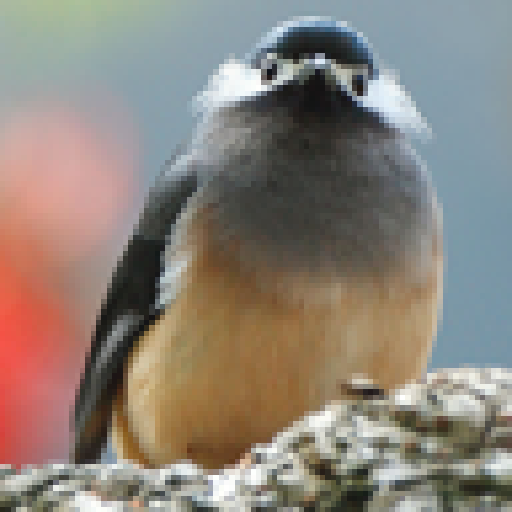}
        \caption{Original}
    \end{subfigure}
    \begin{subfigure}[b]{0.24\linewidth}
    \centering
        \includegraphics[width=\textwidth]{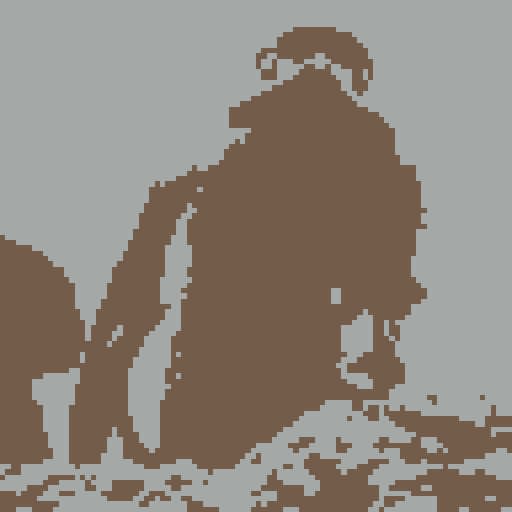}
        \caption{MedianCut}
    \end{subfigure}
    \begin{subfigure}[b]{0.24\linewidth}
    \centering
        \includegraphics[width=\textwidth]{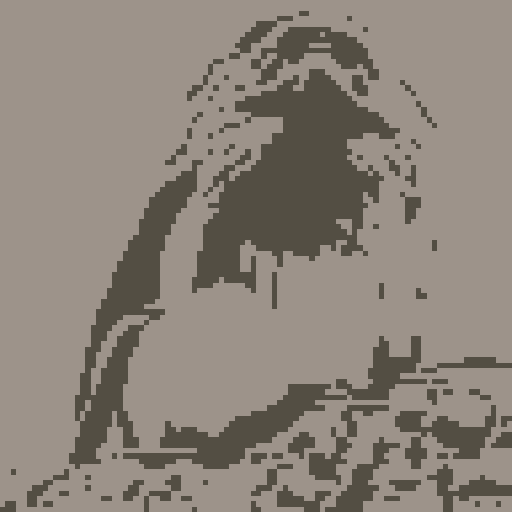}
        \caption{ColorCNN}
    \end{subfigure}
    \begin{subfigure}[b]{0.24\linewidth}
    \centering
        \includegraphics[width=\textwidth]{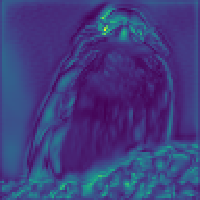}
        \caption{Activation}
    \end{subfigure}
\vspace{-2mm}
\caption{Example of 1-bit color quantization results. 
Visualized based on its auto-encoder output, column (d) shows the critical structures identified by ColorCNN. 
}
\vspace{-3mm}
\label{fig:activation_visualization}
\end{figure}

\begin{figure*}[t]
\begin{center}
\centering
\includegraphics[width=\linewidth]{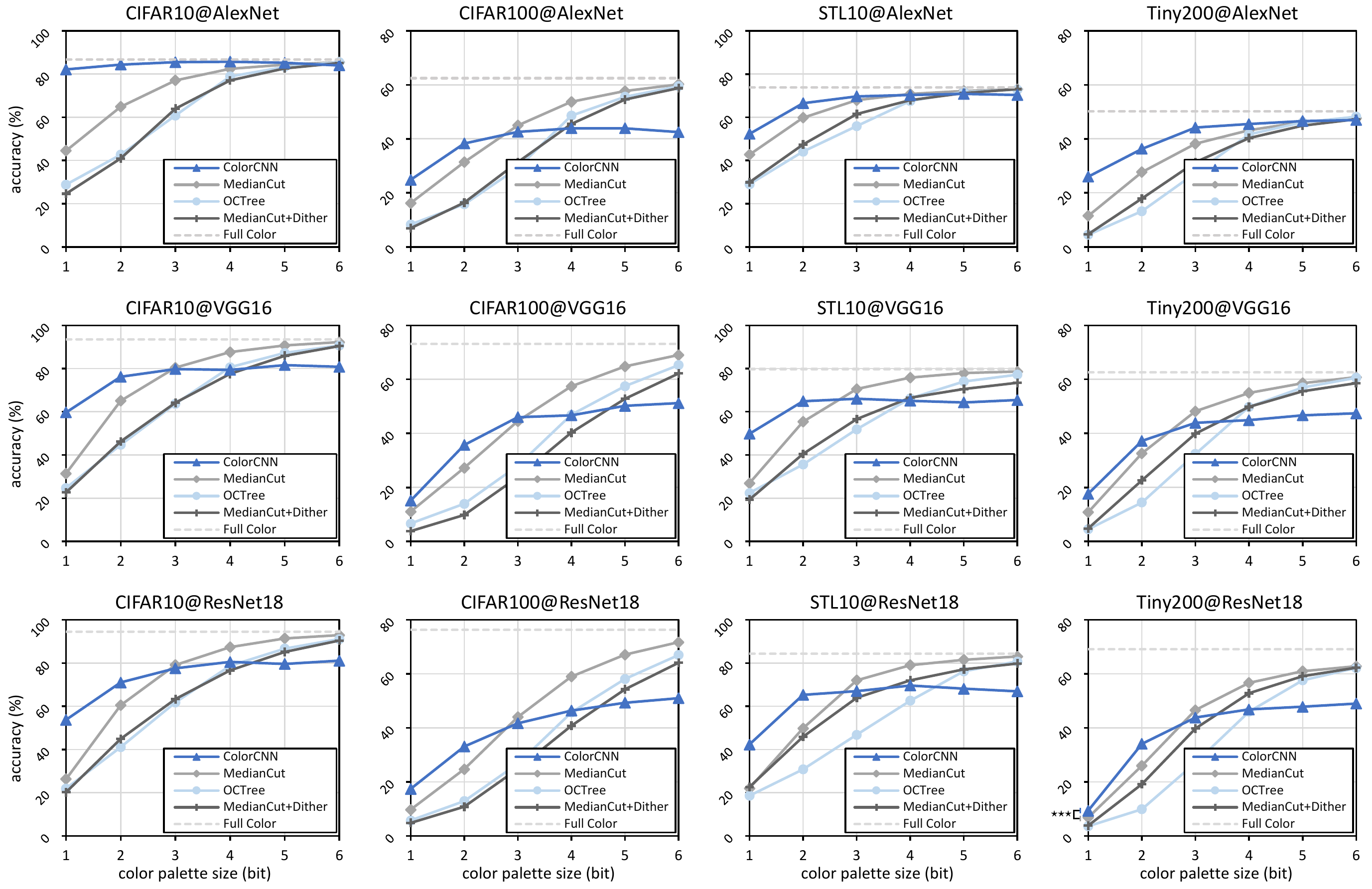}
\vspace{-2mm}
\caption{Top-1 classification accuracy of color quantized images on four datasets with three networks. We observe that ColorCNN is significantly superior to MedianCut, OCTree, and MedianCut+Dither under low-bit quantization. $\star\star\star$ means that the accuracy difference between ColorCNN and MedianCut, is \textbf{statistically very significant} (\ie, $p\text{-value} <0.001$), 
in 1-bit color space, tiny-imagenet-200 dataset, and ResNet18 classifier.
Note that quantization under the large color space is not the focus of this work. More discussion around this consideration is provided in Section~\ref{sec:Discussion}. 
}
\vspace{-5mm}
\label{fig:accuracy_curves}
\end{center}
\end{figure*}

\textbf{Visualization for ColorCNN low-bit quantization.} As shown in Fig.~\ref{fig:activation_visualization}, ColorCNN effectively preserves the shapes, textures and other structures. For instance, airplane wings, vehicle tires and windshield, and bird cheek and belly. In column (\textbf{d}), we find the feature maps extracted via auto-encoder show high activation on the informative edges, details, and textures. We further show the accuracy increase from these critical structures in Fig.~\ref{fig:accuracy_curves}.  

\textbf{Low-bit color quantization performance.} We report top-1 classification accuracy with color quantized images on all four datasets using three networks in Fig.~\ref{fig:accuracy_curves}. We choose MedianCut~\cite{heckbert1982color}, OCTree~\cite{gervautz1988simple}, and MedianCut with dithering~\cite{floyd1976adaptive} for comparisons. 

First, the proposed ColorCNN method brings a consistent and significant improvement over the traditional quantization method in \textbf{a small color space}. Using AlexNet as the classification network, 1-bit quantization results of ColorCNN reach 82.1\%, 24.8\%, 52.3\%, and 26.0\% on four datasets, respectively. This translates into 37.6\%, 8.6\%, 9.6\%, and 14.5\% absolute accuracy increases and 92.9\%, 56.5\%, 66.5\%, and 75.3\% relative accuracy increases over the traditional color quantization methods. Due to more demanding task in the 100-way classification compared to the 10-way one (on same data), the improvements (absolute and relative) are smaller on CIFAR100 compared to CIFAR10. Nonetheless, these non-trivial accuracy increases still prove the effectiveness of the proposed method. 
We point out that classifiers trained on original images 1) have significantly lower performance on low-bit images, 2) but still manage to classify some low-bit images. The color quantized images and original images can be regarded as two different domains. The two domains differ significantly in appearance, but also share the same set of classes and some appearance similarities. Their feature distributions thus are very different but still have some level of overlap, which accounts for both sides of the phenomenon. 

\begin{table}[t!]
\begin{center}
    
\resizebox{\linewidth}{!}{
\begin{tabular}{l|c|c|c|c}
\toprule
         & \multicolumn{1}{c|}{CIFAR10} & \multicolumn{1}{c|}{CIFAR100} & \multicolumn{1}{c|}{STL10} & \multicolumn{1}{c}{Tiny200} \\ \hline
AlexNet~\cite{krizhevsky2012imagenet}  & 86.8                         & 62.5                          & 73.8                       & 50.2                                   \\ \hline
VGG16~\cite{simonyan2014very}    & 93.5                         & 73.1                          & 79.8                       & 62.6                                   \\ \hline
ResNet18~\cite{he2016deep} & 94.6                         & 76.3                         & 84.3                       & 69.1                                   \\ \bottomrule
\end{tabular}
}
\vspace{-1mm}
\caption{Top-1 test accuracy (\%)  of classification networks.}
\vspace{-3mm}
\label{tab:classifier}
\end{center}
\end{table}

Second, ColorCNN is usually inferior to the traditional methods under \textbf{a large color space}. As discussed in Section~\ref{sec:Discussion}, 
ColorCNN does not formulate color quantization as a clustering problem. This will naturally lead to inferior results under a larger number-of-cluster (color space size), since the per-pixel approach in ColorCNN cannot enforce intra-cluster similarity. 

Third, preserving structures via dithering is found not useful. Dithering generates a noise pattern based on the quantization error of the color values. It can further remove the flat color region and false contours. 
Without consideration of the structure and semantic, dithering still fails to preserve the semantic-rich structures in a restricted color space, which further leads to inferior accuracy. 

\textbf{Impact of different classification networks.} We compare the quantization performance with different classification networks in different rows of Fig.~\ref{fig:accuracy_curves}. It is found that stronger classifier has lower accuracy in an extremely small color space. 
A stronger classifier can add more transformation to the image data, extracting more expressive features, thus having higher accuracy. 
However, when the color space is limited, these colors and structures all diminish. This can lead to larger drifts in feature space from the stronger classifiers, since they add more transformation to the input, which ultimately leads to lower accuracy.

We also find that ColorCNN performance is not always higher when trained with stronger classifiers. In fact, stronger classifiers can easily classify the quantized image $\bm{\widetilde{x}}$ during training,  which can lead to earlier, less mature convergence. Given the difference between train-time and test-time forward pass $\widetilde{g_\psi}\!\left(\cdot\right)$ and ${g_\psi}\!\left(\cdot\right)$, this less mature convergence can result in more overfitting and lower accuracy.

\begin{figure}[t]
\centering
\includegraphics[width=0.85\linewidth]{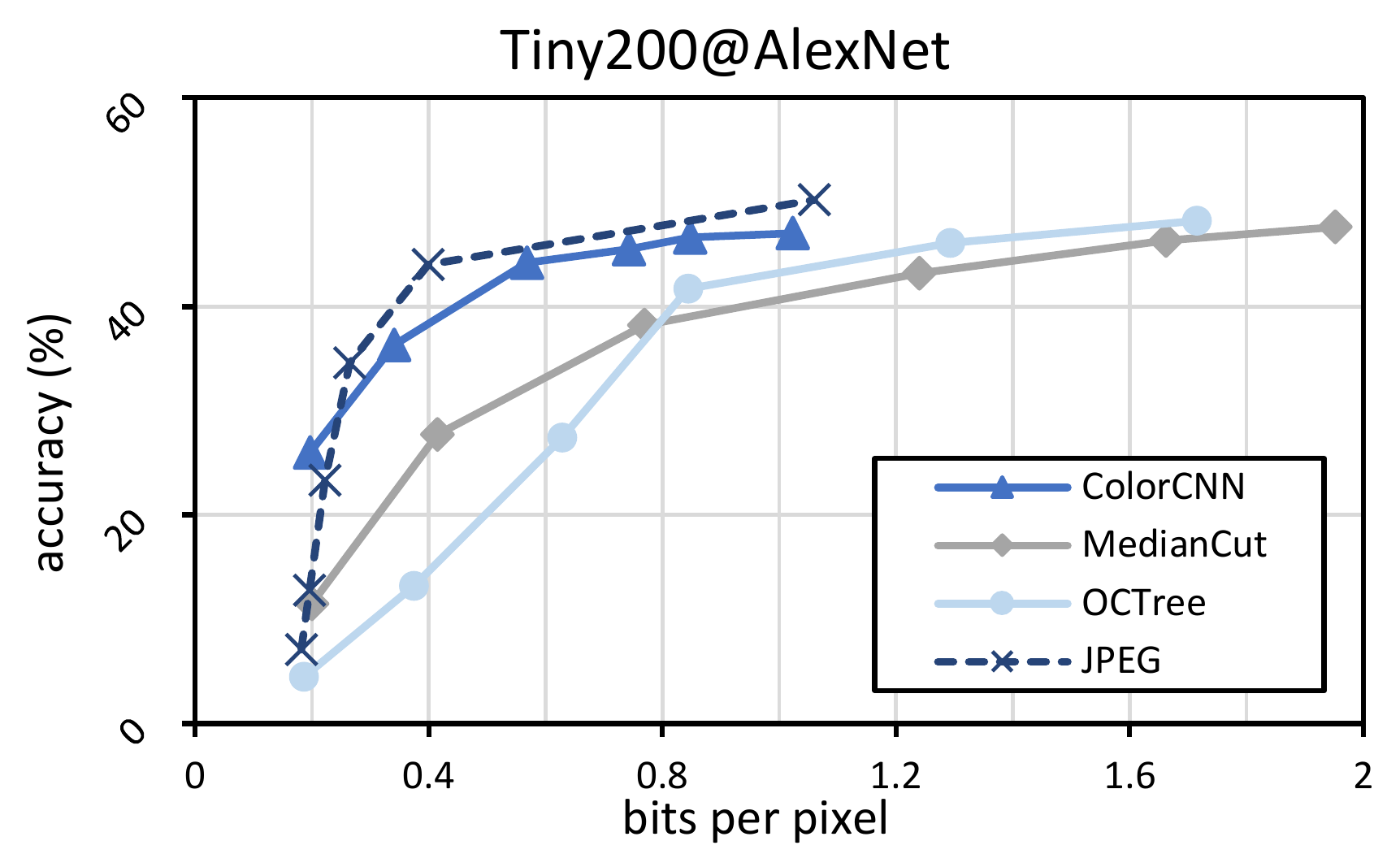}
\vspace{-1mm}
\caption{Classification accuracy under different bit-rate. Solid lines refer to color quantized image encoded via PNG. Dotted line refers to JPEG encoding as reference. 
For color quantization methods, bit-rate from low to high are quantized images under color space size from 1-bit to 6-bit. 
}
\vspace{-3mm}
\label{fig:compression_curves}
\end{figure}

\textbf{Low-bit color quantization as image compression.} In Fig.~\ref{fig:compression_curves}, as the color space size grows from 1-bit to 6-bit, the quantized images take a higher bit-rate when encoded with PNG, and have higher test accuracy. 
When compared to traditional color quantization methods, ColorCNN can reach higher test accuracy under a lower bit-rate. As shown in Fig.~\ref{fig:6-bit}, even if 6-bit color space is allowed, ColorCNN only uses a few colors for the majority of the image. Not using all colors evenly will introduce a lower information entropy~\cite{shannon1948mathematical}, which leads to a smaller bit-rate when compressed losslessly via PNG. This suggests ColorCNN can extract the key structure of an image more effectively. 
Moreover, under 0.2 bits per pixel, 1-bit ColorCNN quantization can even outperform JPEG compression by 13.2\%, which has more than 2 colors. This clearly demonstrates the effectiveness of ColorCNN. 

\subsection{Ablation Study}
We set the regularization weight $\gamma=0$ or color jitter weight $\xi=0$, to see the ColorCNN performance without either of those. Results are shown in Table~\ref{tab:ablation}.

First, we find that removing the regularization term causes an accuracy drop. In fact, without the regularization, fewer colors are chosen during test-time. This is because the softmax color filling during training can introduce more colors in the image, as shown in Fig.~\ref{fig:comparison}. This difference in color filling leads to overfitting and a 2.2\% accuracy drop. 

Second, no color jitter also leads to an accuracy drop. Without color jitter, the train-time quantization can be too easy for the pre-trained classifier. This can lead to premature convergence, which further hurts accuracy by 1.9\%.

Third, higher bit-rate does not necessarily lead to higher accuracy, what matters is preserving the critical structure. Under similar accuracy, ColorCNN without regularization actually has a smaller bit-rate than without color jitter. 
With both regularization and color jitter, the bit-rate becomes even higher. However, this time, since the introduced structures can help the recognition, ColorCNN achieves the highest accuracy. We find this coherent with the compression rate curves in Fig.~\ref{fig:compression_curves}, where ColorCNN achieves higher accuracy under lower bit-rate, since it preserve the more critical structures.

\subsection{Variant Study}
\begin{figure}[t]
\centering
\includegraphics[width=\linewidth]{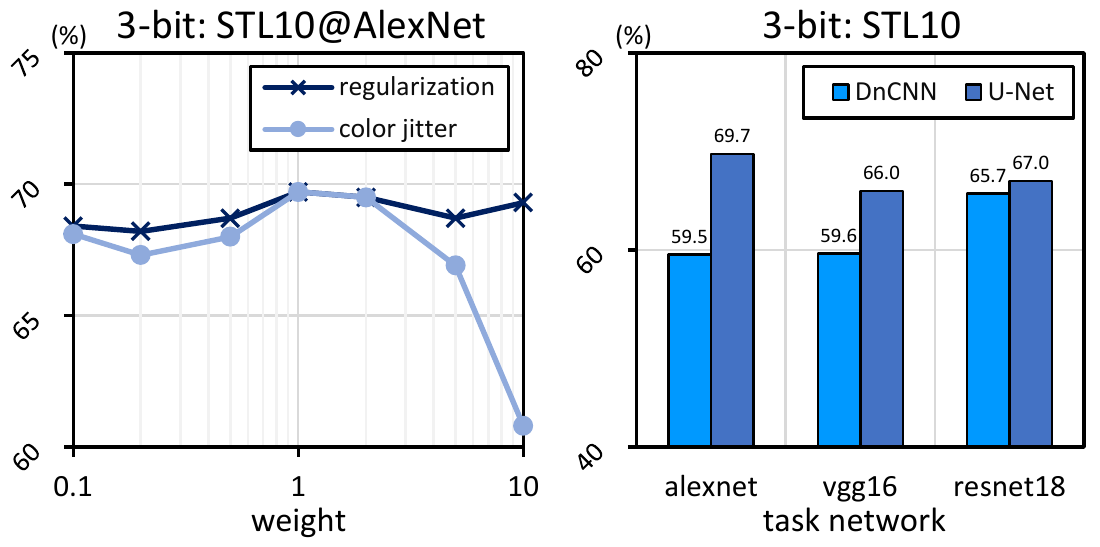}
\vspace{-4mm}
\caption{Performance comparison with different weights and different auto-encoder backbone. 
}
\vspace{-1mm}
\label{fig:variants}
\vspace{-3mm}
\end{figure}

We compare the recognition accuracy of ColorCNN and its variants, including ones with different hyper-parameters, and one with different auto-encoder backbone (Fig.~\ref{fig:variants}). 
ColorCNN performance is lower when the regularization weight is too small or too high. 
Similarly, too small or too large a color jitter can also result in a huge accuracy drop. This is because setting the weight too small or too large leads to either too small a influence, or completely overshadowing anything else. For both regularization and color jitter, we witness the highest accuracy when the weight is set to $1$, which corresponds to our hyper-parameter setting.

When the auto-encoder backbone is replaced with DnCNN~\cite{zhang2017beyond}, the ColorCNN performance degrades under all classification networks. Unlike U-Net, DnCNN does not have bypasses to maintain the local structure. As a result, its quantization results might have structure misalignment, which hurts classification accuracy.

\begin{table}[t]
\resizebox{\linewidth}{!}{
\centering
\begin{tabular}{l|c|c|c}
\toprule
                   & \multicolumn{1}{l|}{Accuracy (\%) } & \multicolumn{1}{l|}{\#color/image } & \multicolumn{1}{l}{\#bit/pixel } \\ \hline
ColorCNN           & \textbf{69.7}                      & \textbf{8.0}                       & 0.425                               \\ \hline
w/o regularization & 67.5                               & 5.1                                & \textbf{0.323}                               \\ \hline
w/o color jitter   & 67.8                               & \textbf{8.0}                               & 0.390                               \\ \bottomrule
\end{tabular}
}
\caption{Test results under 3-bit color quantization, STL10 dataset, AlexNet classifier. }
\label{tab:ablation}
\end{table}

\section{Conclusion}
In this paper, we investigate the scientific problem of keeping informative structures under an extremely small color space. 
In such cases, traditional color quantization methods tend to lose both color and structure, making its output incomprehensible to neural networks. 
In order to maintain the critical structures in the quantized images so that they can be correctly recognized, we design the ColorCNN color quantization network. 
By incorporating multiple cues for a comprehensive quantization decision making, ColorCNN effectively identifies and preserves the informative structures, even in the extreme conditions. An end-to-end training method is also designed for ColorCNN, so as to maximize the performance of the quantized image on the neural network tasks. The effectiveness of the proposed method is demonstrated on four datasets with three classification networks.

\section{Acknowledgements}
Dr. Liang Zheng is the recipient of Australian Research Council Discovery Early Career Award (DE200101283) funded by the Australian Government.

{\small
\bibliographystyle{ieee_fullname}
\bibliography{egbib}
}

\end{document}